\documentclass[12pt,fleqn]{article}
\usepackage{xiiiemc}
\usepackage{natbib}
\usepackage{fancyhdr}
\usepackage{fancyvrb}
\usepackage{color}
\usepackage{wallpaper} 
\usepackage{titlesec}   
\usepackage{float} 	
\usepackage[usenames,dvipsnames]{xcolor}
\usepackage{hyperref} 	
\hypersetup{
	pdfcreator={LaTeX with abnTeX2},
	pdfkeywords={abnt}{latex}{abntex}{USPSC}{trabalho acadêmico}, 
	colorlinks=true,       		
	linkcolor=blue,          	
	citecolor=blue,        		
	filecolor=magenta,      		
	urlcolor=blue,
	allbordercolors=black,
	bookmarksdepth=4
}
\newcommand{\URL}[1]{\url{\detokenize{#1}}}
\usepackage{tocloft}
\titleformat{\section}
  {\normalfont\bfseries}{\thesection.}{0.5em}{}
 
\renewcommand\thesection{\arabic{section}}

\let\OLDthebibliography\thebibliography
\renewcommand\thebibliography[1]{\OLDthebibliography{#1} \setlength{\parskip}{0pt}\setlength{\itemsep}{0pt plus 0.3ex}}

\title{ENHANCEMENTS OF LINKED DATA EXPRESSIVENESS FOR ONTOLOGIES}

\author
    {\rm \begin{tabular}{l} 
    \textbf{Renato Fabbri}$^{1}$ - {\textnormal renato.fabbri@gmail.com}\\%
    {\fontsize{11}{0}\selectfont $^{1}$University of São Paulo, Institute of Mathematical and Computer Sciences - São Carlos, SP, Brazil}\vspace*{-0.05cm} \\
  \end{tabular}}

\fancypagestyle{firspagetstyle}
{
	\lhead{}
	\fancyhead[C]{%
		\includegraphics[width=0.9\linewidth]{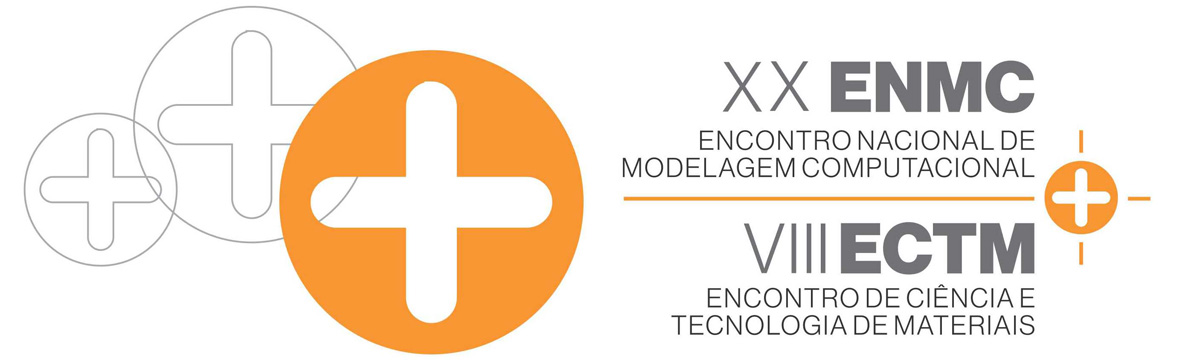}\\%
		{\scriptsize \fontfamily{phv}\fontseries{b}\selectfont \color[rgb]{0.45,0.45,0.45}
		16 a 19 de Outubro de 2017\\
		Instituto Politécnico - Universidade do Estado de Rio de Janeiro\\
		Nova Friburgo - RJ\\
	    }
	}
	\renewcommand{\headrulewidth}{0.0pt}
	\fancyfoot[C]{\footnotesize \parbox{15cm} {\centering  \fontsize{7.5}{0}\selectfont \it Anais do XX ENMC – Encontro Nacional de Modelagem Computacional e VIII ECTM – Encontro de Ciências e Tecnologia de Materiais,  Nova Friburgo, RJ – 16 a 19 Outubro 2017}} 
	\rhead{}
}

\begin{document}
\maketitle

\thispagestyle{firspagetstyle}

\fancyhead[L]{\footnotesize{\fontsize{7.5}{0}\selectfont \it XX ENMC e VIII ECTM\\
	16 a 19 de Outubro de 2017\\
	Instituto Politécnico Universidade do Estado do Rio de Janeiro – Nova Friburgo - RJ\\}}
\renewcommand{\headrulewidth}{0.0pt}
\fancyfoot[C]{\footnotesize \parbox{15cm} {\centering  \fontsize{7.5}{0}\selectfont \it Anais do XX ENMC – Encontro Nacional de Modelagem Computacional e VIII ECTM – Encontro de Ciências e Tecnologia de Materiais,  Nova Friburgo, RJ – 16 a 19 Outubro 2017}} 
\rhead{}

\begin{abstract}
The semantic web has received many contributions of researchers as ontologies which,
in this context, i.e. within RDF linked data, are formalized conceptualizations that might use
different protocols, such as RDFS, OWL DL and OWL FULL.
In this article, we describe new expressive techniques which were found necessary after
elaborating dozens of OWL ontologies for the scientific academy, the State and the civil society.
They consist in: 1) stating possible uses a property might have without incurring into axioms or restrictions;
2) assigning a level of priority for an element (class, property, triple); 3) correct depiction in diagrams
of relations between classes, between individuals which are imperative, and between individuals which are optional;
4) a convenient association between OWL classes and SKOS concepts.
We propose specific rules to accomplish these enhancements and exemplify both its use
and the difficulties that arise because these techniques are currently not established as standards to the ontology designer.  
\end{abstract}


\pagestyle{fancy}

\section{INTRODUCTION}
The semantic web is constituted by data which is linked in the same way
web pages are: through HTTP and URLs.
W3C recommendations are the main sources of protocols and best practices.
In practice, the terms 'linked data' and 'semantic web' are most often used interchangeably.
A distinction might arise in some contexts where one needs to refer to the linked data or
the semantic web created by all or some portion of linked data, but, as a knowledge and technological
field, the terms are equivalent.

The semantic web is built using the Resource Description Framework (RDF).
The RDF data model is based on making statements in the form of triples
(``subject-predicate-object'') and using Unique Resource Identifiers (URIs) for
objects and concepts.
It is also part of the framework to use URIs that are URLs whenever possible,
to enable the data linkage.
Accordingly, one can write:
\begin{Verbatim}[fontsize=\footnotesize,commandchars=\\\{\}]
<\URL{http://example.org/people/mary}>
	<\URL{http://example.org/properties/name}> "Mary Shastacian" .
<\URL{http://example.org/people/mary}>
	<\URL{http://example.org/properties/age}> "57" .
<\URL{http://example.org/people/mary}> 
	<\URL{http://example.org/properties/likes}> 
		<\URL{http://example.org/concepts/reading}> .
\end{Verbatim}
\noindent to express that there is person called Mary Shastacian which is 57 years old and likes reading.
There are many formats to write/serialize RDF data.
The example above is written in Turtle and we will use this format throughout this document.

One of the most important notions of the semantic web is that of an ontology.
An ontology, in this context, is a formalized conceptualization, comprised by
concepts and relations between the concepts and between the relations themselves.
In current semantic web, most simple ontologies are written using the RDFS protocol,
by which one can specify, among other things:
\begin{itemize}
	\item concepts;
	\item properties, which are concepts that are used as predicates and thus relate concepts;
	\item special relations between concepts that state that one concept is more general than the other\footnote{This
		kind of relation is called hypernymy. Examples: mammal is a hypernym of monkey, drink is a hypernym of beer.};
	\item the subjects and objects that can occurs in a triple where a specific property is a predicate.
\end{itemize}

One can also write an ontology by using OWL,
in which all the expressive capabilities of RDFS are
available, but one can also, among other things:
\begin{itemize}
	\item state ``property axioms'', i.e. specify if a property is e.g. reflexive or transitive;
	\item state ``class restrictions'', i.e. specify if a class instance e.g. necessarily holds a relation to another class instance or data.
\end{itemize}
OWL has a richer vocabulary than RDFS and is (way more) complex.
This complexity is a drawback together with the greater computational cost
for performing inference.
The advantage is the greater power to represent conceptualizations.

The semantic web practices evolves by constant practice and recommendations written
by many specialists usually after years of research and discussion.
Therefore, this document is written in a very humble manner,
without the pretense of yielding new recommendations abruptly,
but as a way to report issues and candidate workarounds/solutions
that emerged by years of intense practice in designing ontologies~\citep{pnud5,fabbri1,losd,ops}.

Section~\ref{sec:prob} is dedicated to exposing some of the problems
in current RDFS and OWL ontologies, with emphasis on the ones
to which this document holds proposals to resolve or lessen.
In Section~\ref{sec:prop} such proposals are described and discussed.
Final remarks are articulated in Section~\ref{sec:con}.

\section{THE PROBLEMS}\label{sec:prob}
The semantic web technology standards are very appreciated by the researchers, developers and users~\citep{semApr}.
Even so, there are many issues regarding current state of the art.
For example, if an individual of a class is found also to be a class,
it implies OWL Full which is a setting where inferences are presently regarded as not computationally feasible.
This document is dedicated to proposals that resolve (or alleviate)
the problems described in the following subsections.

\subsection{No (standard) way to express potential relations}\label{sec:rel}
There is no recommendation about how to bind to a property a \emph{possible}
subject or object.
Using \texttt{rdfs:domain} or \texttt{rdfs:range} has the consequence that
\textbf{any} subject or object (respectively) will be inferred to be a member of the class specified by
range or domain.
There are, however, many situations where this is not intended.
For example, one might find or create a property \texttt{example:name}.
Suppose it is intended to be used to express names of animals,
say because the person created the property while designing an ontology about animals.
If that is formalized in the property by \texttt{rdfs:domain},
it implies that whenever the \texttt{example:name} is used, the subject
will be inferred (e.g. by an automated reasoner) to be an animal.
Now suppose someone is describing a tool in RDF and wants to write its name.
If he/she uses the \texttt{example:name} property, it will be erroneously
expressing that the tool is an animal.
Such a person is required to create another property \texttt{example2:name}
to write the name(s)\footnote{The \texttt{rdfs:label} is often used to assign names,
but the example is simple and valid.
It is straightforward to 1) conceive some property to which a potential domain might be intended
but the restriction implied by \texttt{rdfs:domain} is not convenient;
or 2) find an example in a widely adopted ontology, e.g. \texttt{foaf:topic}
has \texttt{rdfs:domain foaf:Document}.}.
What is worse: if the person wants to create an ontology about tools,
and wants to express that tools are likely to have names,
the way to represent this if by assigning to the property \texttt{exampleX:name} the ``domain tool''
and fall into same problem
as above, but now with tools instead of animals.
If \textbf{all} the tools have, without exception, a name, then one can express
it as an OWL existential class restriction.
But, again, that is often not the case. Example: algorithms are likely to have names,
but not all of them have names.
One workaround to this problem is to describe these potential relations as notes to the
classes and properties, but it does not favor the machine readable design of the semantic web.

\subsection{All classes and properties are equally fundamental}\label{sec:fun}
The relevance of this issue can be grasped by
tackling the task of expressing a conceptualization for humans.
Say someone wrote an ontology about machine learning and that it is very complete
in the sense that it contains many concepts, all those that the author could find.
It is straightforward to acknowledge that it will be very difficult for a newcomer
to make sense of the ontology.
Also, in many cases it is difficult for anyone to make sense of an ontology when it
comprises many concepts and relations.
Thus, it is very useful to have a standard way to express that some set of
concepts or triples are more fundamental then others in a layered fashion,
but there isn't a standard way to achieve this.

\subsection{Distinction in diagrams of relations between classes and individuals and between imperative and optional relations}\label{sec:dia}
This is not a problem found in formalizing the ontology,
but a problem that arises when depicting the ontology as a diagram.
For example, the (class of) dinosaurs were succeeded by the (class of) mammals,
this is a relation between classes.
A specific animal might have siblings, this is a relation between individuals.
Also, a relation might be imperative for a class (i.e. it might be expressed as a class restriction)
or optional.
These are all poorly distinguished in diagrams that are meant to represent ontologies.

\subsection{No obvious way to associate an OWL ontology and a SKOS vocabulary}
In a SKOS vocabulary, a concept is an individual of the class \texttt{skos:Concept}.
In an OWL ontology, a concept is a class.
As stated in the beginning of this section, 
if an individual of a class is also a class,
one has an OWL Full context, which is
currently considered to entail inference that is not computable.
This implies that one has to choose between making
an OWL ontology or a SKOS vocabulary or find a way to
make them compatible.
This might be the most complex issue tackled in this document.

\section{ENHANCEMENT PROPOSALS}\label{sec:prop}
As the problems stated in the previous section are distinct,
the proposed enhancements are described case-by-case.

\subsection{Potential domain, range, predicate and post predicate}
To address the problem described in Section~\ref{sec:rel},
I propose creating a new set of properties:
\begin{itemize}
	\item potential/nonrestrictive domain and range of a property might be expressed using\\
		\texttt{example:potentialDomain} and \texttt{example:potentialRange}.
	\item potential predicates when an instance of a class is a subject or an object might
		be expressed using\\
		\texttt{example:potentialPredicate} and \texttt{example:potentialPostPredicate}.
\end{itemize}

The \texttt{example:} in these triples should be replaced by a true namespace prefix, such
as \texttt{http://purl.org/socialparticipation/rdfe/}.
This was not done until the moment because we chose to wait,
present this article and distribute it online to find out if the problem is sound, if there is already some
solution currently in use or if there is some trustworthy maintainer that is willing
to keep these properties online in a more suitable namespace.
As a last resort, we (me and partners) might create a reasonable namespace
and bind these properties.

\subsection{Ascribing numeric values that reflect how fundamental is something}\label{sec:props}
To address the problem described in Section~\ref{sec:fun},
every property and class might be a subject to a triple:
\begin{Verbatim}[fontsize=\footnotesize]
	example:something example:fundamental 4 .
\end{Verbatim}
\noindent with ``4'' replaced by any other numerical value.
At first, it seems reasonable to assume that these values are
valid within the ontology and relative to each other.
Accordingly, one might ascribe \texttt{fundamental 100}
some to some nucleus of meaning and lower values to other layers of
concepts and statements with a (more-or-less) equal weight with respect
to being fundamental to the ontology.
As statements are usually not given a URI explicitly
in RDF files (one has to use ``RDF reification'' to give a triple a URI),
to bind a triple or a set of triples to a level,
one will probably be better of by using named graphs.
Reification is a solution, but a rather verbose and unusual one.

\subsection{Different arrow heads and tails}
To address the problem described in Section~\ref{sec:dia},
we notice that: 1) the edges are often small, so using line styles
(such as dashed, dotted or full) is likely to yield bad results often.
The use of colored edges is an option, but given that many
journals require images to be black and white, it is not suitable
as the standard at least at the moment.
Nevertheless notice that different arrow heads are in fact used (almost as a standard) 
as visual cues
to differentiate relations that are yield by properties from those
that associate a class to its superclass,
This is an indicative that using different arrow heads might
be a good strategy to reaching a good solution.
Also, the start of the edge might be a good visual cue to
discern different types of relations.
We should recall here that in OWL a property might be specified to
satisfy a number of axioms, which are also prone to be represented
with these same artifices.

\subsection{OWL class as subclass of SKOS concept}\label{owlSkos}
There are different
strategies for making an OWL ontology and a SKOS vocabulary compatible~\citep{owlSkos}.
The solutions that are fit for general purposes, among the ones that were found or conceived by the author, are:
\begin{itemize}
	\item describe the ontology as OWL then translate it to SKOS (and probably automate the process by means of a script).
	\item State that each OWL class is also a subclass of the \texttt{skos:Concept}.
		This yields a less redundant result, but it might not be in accordance to SKOS and OWL specifications
		or entail OWL Full.
		For example, the use of some SKOS properties might imply that the subject is an individual of the \texttt{skos:Concept} class.
		This solution needs to be further studied and discussed with specialized groups in order to be used with confidence.
\end{itemize}

The problem with the other solutions:
\begin{itemize}
	\item yields OWL Full (if a SKOS concept, an instance, is also an OWL class); or
	\item yields annotations on instances of OWL classes, not on the classes
		(if we annotate instances of OWL classes with SKOS properties); or
	\item yields a less formal ontology (if we derive the ontology from the vocabulary); or
	\item extrapolates the purpose of SKOS
		(giving enough expressive power to SKOS to achieve an OWL without degeneration of the conceptualization).
\end{itemize}

\section{CONCLUSIONS AND FURTHER WORK}\label{sec:con}
This article is a formalization of the issues to allow
the author and collaborators to confirm (or refute) that these problems
and the solutions herein proposed are sound and have not yet received
solutions by the semantic web community.
These issues and solutions are the result of years of dedication
to the subject, including learning by (online and in-person) classes
and designing dozens of OWL ontologies for academic and State purposes~\citep{pnud5,fabbri1,losd}.

Next steps should include:
\begin{itemize}
	\item the pursue of feedback through circulation of this document (e.g. in online forums)
		to enhance the solutions herein proposed and better understand their possible uses (e.g. in reasoners).
	\item The use of these solutions in ontologies, both new ones and others that are already available.
	\item If these issues prove themselves relevant over time, we will consider contributing to the full realization of solutions.
		For example, we might help achieving and maintaining suitable URIs for the properties described in Section~\ref{sec:props}
		or help in the elaboration of convenient arrow heads for the problem described in Section~\ref{sec:dia}.
	\item The confirmation or refutation of the second possible solution to make OWL and SKOS compatible as described in Section~\ref{owlSkos}.
\end{itemize}

\subsection*{\textit{Acknowledgements}}
The author thanks CNPq for the funding received while researching the topic of this article,
the researchers of IFSC/USP and ICMC/USP for the recurrent collaboration in every situation
where we needed directions for investigation.








\end{document}